\documentclass{article}
\usepackage{arxiv}
\usepackage[square,sort,comma,numbers]{natbib}
\usepackage[utf8]{inputenc} 
\usepackage[T1]{fontenc}    
\usepackage[shortlabels]{enumitem}
\usepackage[obeyspaces]{url}
\usepackage{wrapfig}
\usepackage{booktabs}       
\usepackage{amsfonts}       
\usepackage{nicefrac}       
\usepackage{microtype}      
\usepackage{lipsum}		
\usepackage{graphicx}
\usepackage{natbib}
\usepackage{doi}
\usepackage{multirow}
\usepackage{tabularx}
\usepackage{tabulary}
\usepackage{makecell}
\usepackage{tabularx,ragged2e}
\usepackage{fancyhdr, blindtext}
\usepackage{subfigure}
\usepackage{float}
\usepackage{hyperref}
\usepackage{hhline}
\usepackage{array}
\usepackage{amsmath}
\usepackage{diagbox}
\usepackage{color}
\usepackage{textcomp}
\usepackage{comment}
\usepackage{relsize}
\usepackage[symbol]{footmisc}
\usepackage{caption}
\usepackage{chngcntr}

\setcitestyle{square}
\newcommand{\triplet}[1]{\textlangle{\textit{#1}}\textrangle{}}

\title{Data Splits and Metrics for Method Benchmarking on Surgical Action Triplet Datasets}

\date{April 12, 2022}

\author{ 
\href{http://orcid.org/0000-0003-4777-0857}{\includegraphics[scale=0.06]{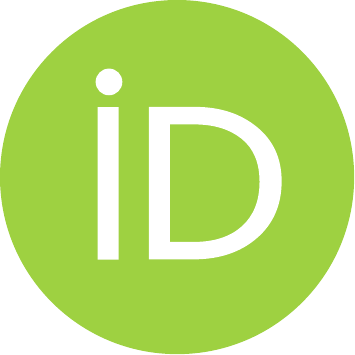}\hspace{1mm}Chinedu Innocent Nwoye}\\
	ICube Laboratory, CNRS,\\
	University of Strasbourg, France \\
	\texttt{nwoye@unistra.fr} \\
	\And
	\href{https://orcid.org/0000-0002-5010-4137}{\includegraphics[scale=0.06]{orcid.pdf}\hspace{1mm}Nicolas Padoy} \\
	IHU Strasbourg, France\\
	ICube, University of Strasbourg, CNRS, France\\
	\texttt{npadoy@unistra.fr} \\
}

\renewcommand{\headeright}{A Benchmark Study}
\renewcommand{\undertitle}{A Benchmark Study}
\renewcommand{\shorttitle}{CholecT45 \& CholecT50 Dataset Splits and Metrics}

\hypersetup{
pdftitle={CholecT50 Dataset Splits and Metrics},
pdfsubject={q-bio.NC, q-bio.QM},
pdfauthor={Chinedu Innocent Nwoye, Nicolas Padoy},
pdfkeywords={Surgical activity recognition, Tool-tissue interaction, Action triplet, CholecT40, CholecT45, CholecT50},
}

\begin{document}
\maketitle

\begin{abstract}
In addition to generating data and annotations, devising sensible data splitting strategies and evaluation metrics is essential for the creation of a benchmark dataset. This practice ensures consensus on the usage of the data, homogeneous assessment, and uniform comparison of research methods on the dataset.
This study focuses on CholecT50, which is a 50 video surgical dataset that formalizes surgical activities as triplets of \triplet{instrument, verb, target}.
In this paper, we introduce the standard splits for the CholecT50 and CholecT45 datasets and show how they compare with existing use of the dataset. CholecT45 is the first public release of 45 videos of CholecT50 dataset.
We also develop a metrics library, \path{ivtmetrics}, for model evaluation on surgical triplets.
Furthermore, we conduct a benchmark study by reproducing baseline methods in the most predominantly used deep learning frameworks (PyTorch and TensorFlow) to evaluate them using the proposed data splits and metrics and release them publicly to support future research.
The proposed data splits and evaluation metrics will enable global tracking of research progress on the dataset and facilitate optimal model selection for further deployment.
\end{abstract}
\keywords{: Surgical activity recognition \and Tool-tissue interaction \and Action triplet \and CholecT40 \and CholecT45 \and CholecT50\\[0.05in]}



\begin{figure}[ht]
  \begin{center}
    \includegraphics[width=\textwidth]{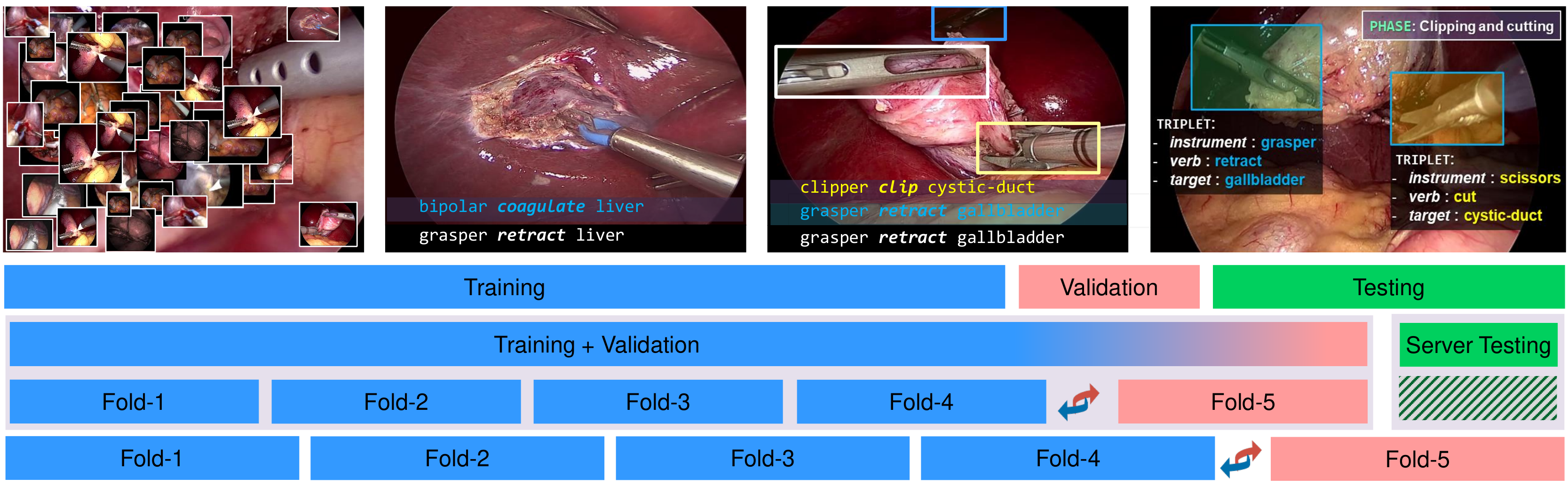}
    \caption{An illustration of the  dataset splits. {\it First row:}  CholecT50 split as used in the Rendezvous \cite{nwoye2021rendezvous}. {\it Second row:} CholecT50 split as used in the CholecTriplet challenges \cite{nwoye2022cholectriplet2021,nwoye2023cholectriplet2022}. {\it Third row:} the official cross-validation split on CholecT45. {\it Fourth row:} the official cross-validation split on CholecT50.}
    \label{fig:datasplits}
  \end{center}
\end{figure}

\section{Introduction}
The use of Artificial Intelligence (AI) techniques is increasingly driving research and development across many disciplines. Yet, there has been a delay in introducing large-scale data science to interventional medicine, partly due to the unavailability of large annotated datasets \cite{maier2017surgical}.
While huge efforts have been made in creating small to medium scale datasets \cite{sznitman2012data,twinanda2016endonet,jin2018tool,al2019cataracts,grammatikopoulou2019cadis,nwoye2020recognition,ross2020robust,bawa2021saras,nwoye2021rendezvous}, little or no effort has been made to standardize the data usage for tracking the global research progress.
For instance, in laparoscopic and cataract surgeries, many published methods on the most prominent tasks of surgical phase \cite{twinanda2016endonet,dergachyova2016automatic,gao2021trans,funke2018temporal} and tool detection \cite{garcia2017toolnet,vardazaryan2018weakly,nwoye2019weakly} are reported on varying data splits of the same dataset, e.g: Cholec80 \cite{twinanda2016endonet}, Cataract \cite{al2019cataracts}, etc.
Without a consensus data split, tracking research progress on these experimental datasets is not straightforward.
Oftentimes, it complicates results comparison, making model selection for further clinical translation more challenging.

In this paper, we present standard data splits for the recently introduced CholecT50 dataset \cite{nwoye2021rendezvous}. The label formalism in the dataset provides comprehensive and fine-grained details on every tool-tissue interaction in any given surgical scene. 
A subset of the dataset, named {\it CholecT45}, was released after the CholecTriplet2021 challenge \cite{nwoye2022cholectriplet2021} while withholding 5 test set videos from public access.
The remaining part of the dataset is planned to be released after the CholecTriplet2022 challenge.

The data split patterns, illustrated in Fig. \ref{fig:datasplits}, are fashioned on three criteria:
\begin{enumerate}
    \item {\it Reproducibility}: to maintain consistent splits with the earlier published experiments that introduced the dataset,
    \item {\it Accessibility}: owing that the dataset is gradually released in batches over time, we consider a representative setup for its utilization and fair comparison,
    \item {\it Thoroughness}: to counter the effect of class-imbalance predominant in a single test set by using a rigorous and exhaustive $k$-fold cross-validation approach, which enables alternating evaluation on the entire dataset.
\end{enumerate}

Furthermore, we define and standardize the evaluation metrics for assessing the quality of triplet detection and recognition on the dataset. These metrics build on the evaluation setup used in earlier research \cite{nwoye2020recognition,nwoye2021deep} on the dataset. In this work, we describe the evaluation algorithm and develop a standard metrics library, {named \it ivtmetrics}, for both triplet recognition and detection/localization evaluation.
The metrics library is available online and can be installed via \verb|pip| or \verb|conda| package installers for method development and validation. The metrics library is usable in all python-based deep learning frameworks.

Finally, we re-implement our previously proposed deep learning methods for surgical action triplet recognition in two widely used deep learning frameworks: PyTorch and TensorFlow.
The reproduced models are evaluated on the proposed data splits for CholecT45 and CholecT50 using the developed {\it ivtmetrics} library thus providing baselines for future comparison.
By conducting this benchmark experiment on the newly introduced dataset, this study provides a definition of standard practice for the official data splits and an evaluation protocol to guide future research.

The CholecT45 and CholecT50 datasets are released on \url{http://camma.u-strasbg.fr/datasets}. The evaluation metrics library is installable using python package and environment managers (pip and conda). The code and weights for the reproduced models are available on the CAMMA public GitHub \url{https://github.com/CAMMA-public}.

In the following sections, we present the proposed data splits and their constituting videos, followed by the evaluation protocols and the developed metrics library. Afterwards, we present the reproduced models, their performance across the proposed splits, and comprehensive per-category performance analysis on the two datasets. 

\section{Data Splits}
\label{sec:dataset}
CholecT50 \cite{nwoye2021rendezvous} is a surgical dataset for action triplet recognition. It is an extension of CholecT40 dataset \cite{nwoye2020recognition} with additional $10$ videos. It contains 50 videos of laparoscopic cholecystectomy surgery annotated with 100 action triplet classes. 
Action triplet is a formalism  to represent fine-grained activity in the form of \triplet{instrument, verb, target}. In this dataset, they are composed from $6$ instruments, $10$ verbs, and $15$ target classes resulting in over $151$K triplet instances at $1$ fps video frames. 

In the CholecTriplet2021 \cite{nwoye2022cholectriplet2021} / CholecTriplet2022 \cite{nwoye2023cholectriplet2022} challenges, the participants are given access to a subset of the CholecT50 dataset, also known as CholecT45. This subset is the first public release of the CholecT50 dataset available on \url{http://camma.u-strasbg.fr/datasets}.
The videos of CholecT45 are part of the Cholec80 \cite{twinanda2016endonet} dataset. The remaining videos of CholecT50 are part of the Cholec120 - a superset of Cholec80 dataset. The video indexes correspond between the two datasets with the prefix "video" in Cholec80/Cholec120 changed to "VID" in CholecT45/CholecT50. 
The statistics of the datasets are presented in Table \ref{table:stat}

\begin{table}[ht]    
 \centering
    \setlength{\tabcolsep}{8pt}   \captionsetup{skip=0pt,singlelinecheck=off,justification=raggedright}
    \caption{Statistics of CholecT45 and CholecT50 Datasets.}
    \label{table:stat} 
    \resizebox{\textwidth}{!}{%
    \begin{tabular}{@{}lcccccccccc@{}}
        \toprule
        & \multicolumn{4}{c}{Instance count} && \multicolumn{5}{c}{Category count} \\
        \cmidrule{2-5}\cmidrule{7-11}
        Version & \# Videos & \# Frames & \# Instances & \# Bboxes && Triplets & Instruments & Verbs & Targets & Phases \\ \midrule
        CholecT45 & 45 & 90.5K & 137.9K &--&& 100 & 6 & 10 & 15 & 7 \\
        CholecT50 & 50 & 100.9K & 151.0K & 13.0K && 100 & 6 & 10 & 15 & 7 \\
         \bottomrule
    \end{tabular}
    }
\end{table}

To describe the official usage of CholecT45 and CholecT50 datasets, we present the different splits of the datasets in the following sections.
We first present the data split of CholecT50 as used (a) in the Rendezvous paper \cite{nwoye2021rendezvous} that introduces the dataset and (b) in the CholecTriplet challenges \cite{nwoye2022cholectriplet2021} for reproducibility. Afterward, we present the official cross-validation splits of the CholecT45 and CholecT50 datasets.

\begin{table}[t]
\centering
    \setlength{\tabcolsep}{9pt}
    \captionsetup{skip=0pt,singlelinecheck=off,justification=raggedright}
    \caption{CholecT50 dataset split as used in Rendezvous publication \cite{nwoye2021rendezvous}.}
    \label{table:split:rdv}
    \resizebox{\textwidth}{!}{%
        \begin{tabular}{@{}cccccclccccl@{}}
            \toprule
            \multicolumn{7}{c}{Training} & \phantom{abc} & {Validation} & \phantom{abc} & \multicolumn{2}{c}{Testing} \\ 
            \midrule
            VID01 & VID15 & VID26 & VID40 & VID52 & VID65 & VID79  && VID08 && VID06 & VID51 \\
            VID02 & VID18 & VID27 & VID43 & VID56 & VID66 & VID92  && VID12 && VID10 & VID73 \\
            VID04 & VID22 & VID31 & VID47 & VID57 & VID68 & VID96  && VID29 && VID14 & VID74 \\ 
            VID05 & VID23 & VID35 & VID48 & VID60 & VID70 & VID103 && VID50 && VID32 & VID80 \\
            VID13 & VID25 & VID36 & VID49 & VID62 & VID75 & VID110 && VID78 && VID42 & VID111 \\
            \bottomrule      
        \end{tabular}
    }
\end{table}

\subsection{Rendezvous (RDV) Split of CholecT50 dataset}
This split is used in the original paper \cite{nwoye2021rendezvous} that introduces the dataset. It is presented for reproducibility of the earlier published methods on this task.
In this setup, the dataset is split into three: (1) training, (2) validation, and (3) testing as presented in Table \ref{table:split:rdv}.
The videos in each dataset split are distributed in the same ratio to include annotations from each of the (surgeon) annotators. 
This helps to minimize the effect of annotation bias on the learning algorithm.

\subsection{CholecTriplet Challenge Split of CholecT50 dataset}
This split is introduced by the organizers of the CholecTriplet2021 \cite{nwoye2022cholectriplet2021} and CholecTriplet2022 \cite{nwoye2023cholectriplet2022} challenges for surgical action triplet recognition and detection. Here, it is selected for consistency with the methods presented at the MICCAI 2021 EndoVis challenge \cite{speidel2021endoscopic}.
The dataset is split into two: (1) trainval, and (2) testing set, as presented in Table \ref{table:split:ctp}.
During the challenge and for model hyper-parameter tuning, participants are allowed to further split the trainval split into training and validation subsets at their own discretion. All the videos in the trainval are drawn from the publicly available Cholec80 \cite{twinanda2016endonet} dataset. Nevertheless, the testing set containing 5 videos are not in the public domain.
The rationale for this data split is to ensure that the participants do not have access to the testing set of the challenge dataset for fairness in the competition.
\begin{table}[t]
\centering
    \setlength{\tabcolsep}{11pt}
    \captionsetup{skip=0pt,singlelinecheck=off,justification=raggedright}
    \caption{CholecT50 dataset split as used in CholecTriplet2021 \cite{nwoye2022cholectriplet2021} \& CholecTriplet2022 challenges.}
    \label{table:split:ctp}
    \resizebox{\textwidth}{!}{%
        \begin{tabular}{@{}cccccccccrl@{}}
        \toprule
        \multicolumn{9}{c}{Trainval (=~{\it CholecT45})} & \phantom{abc} & {Testing} \\ \cmidrule{1-9} \cmidrule{11-11}
        VID01 & VID10 & VID22 & VID29 & VID42 & VID50 & VID60 & VID73 & VID05 && VID92\\
        VID02 & VID12 & VID23 & VID31 & VID43 & VID51 & VID62 & VID75 & VID18 && VID96\\
        VID04 & VID13 & VID25 & VID32 & VID47 & VID52 & VID66 & VID78 & VID36 && VID103\\
        VID06 & VID14 & VID26 & VID35 & VID48 & VID56 & VID68 & VID79 & VID65 && VID110\\
        VID08 & VID15 & VID27 & VID40 & VID49 & VID57 & VID70 & VID80 & VID74 && VID111\\
        \bottomrule
        \end{tabular}
    }
\end{table}

\subsection{Official Cross-Validation (CV) Splits for CholecT45 and CholecT50 datasets}
$K$-fold cross-validation is known for its robustness in result analysis.
As some of the triplet classes can be unrepresented in any testing set sampling, cross-validation is a more robust, and stable way of assessing the quality of model predictions on all observed triplet classes.
This enables a result analysis that covers all the 100 class labels of the triplet datasets.
In this setup, the dataset is split into 5 equal subsets called \textit{folds}. Different copies of a model are trained on different combinations of 4 out of 5 folds, each time leaving out one alternating fold for testing.
The final result is averaged over the 5 hold-out testing splits.

To ensure that all folds have similar levels of complexity, the 50 videos of CholecT50 are sorted by their difficulty, determined by procedure duration. The sorted videos are divided into 10 clusters with each containing 5 videos of the same/similar complexity or duration. The videos in each cluster are randomly distributed to all the 5 folds split.

The \textbf{CholecT50 CV split} contains the full 50 video dataset divided into 5 folds with each fold containing 10 videos as presented in Table  \ref{table:split:cv} (rows 1-10).

The \textbf{CholecT45 CV split}, on the other hand,  contains 45 videos of the dataset divided into 5 folds with each fold containing 9 videos each as shown in Table \ref{table:split:cv} (rows 1-9). The CholecT45 excludes the test videos (row 10) of the CholecTriplet challenge. Hence, this split equally supports exhaustive cross-validation but only on the publicly released subset of the entire dataset. 

We recommend the use of the cross-validation splits for research purpose as they allow for a complete evaluation of the 100 triplet classes in CholecT45 and CholecT50 datasets.

\begin{table}[t]
\centering
    \setlength{\tabcolsep}{13pt}
    \captionsetup{skip=0pt,singlelinecheck=off,justification=raggedright}
    \caption{Official cross-validation data splits of CholecT45 and CholecT50 datasets (Recommended for research use).}
    \label{table:split:cv}
    \resizebox{\textwidth}{!}{%
        \begin{tabular}{@{}lllccccccccc@{}}
            \toprule
            &&& {Fold 1} & \phantom{abc} & {Fold 2} & \phantom{abc} &{Fold 3} & \phantom{abc} &{Fold 4} & \phantom{abc} &{Fold 5} \\ 
            \midrule
            1 &
            \multirow{10}{*}{\rotatebox[origin=tr]{90}{CholecT50 Cross-Val. Split}} &
            \multirow{9}{*}{\rotatebox[origin=l]{90}{CholecT45 CV Split}} & 
                  VID79 && VID80 && VID31 && VID42 && VID78\\ 
            2 &&& VID02 && VID32 && VID57 && VID29 && VID43\\ 
            3 &&& VID51 && VID05 && VID36 && VID60 && VID62\\ 
            4 &&& VID06 && VID15 && VID18 && VID27 && VID35\\ 
            5 &&& VID25 && VID40 && VID52 && VID65 && VID74\\ 
            6 &&& VID14 && VID47 && VID68 && VID75 && VID01\\ 
            7 &&& VID66 && VID26 && VID10 && VID22 && VID56\\ 
            8 &&& VID23 && VID48 && VID08 && VID49 && VID04\\ 
            9 &&& VID50 && VID70 && VID73 && VID12 && VID13\\ \cmidrule{3-12}
            10&\multicolumn{2}{c}{}& VID111 && VID96 && VID103 && VID110 && VID92\\
            \bottomrule      
        \end{tabular}
    }
\end{table}

\section{Metrics}
This section describes the metrics and library for surgical action triplet task evaluation.

\subsection{Recognition Average Precision}
Triplet recognition performance is evaluated using the Average Precision (AP) metric measured as the area under the precision-recall ({\it p-r}) curve per class:
\begin{equation}
    \label{metrics:AP}
    AP = \int_{0}^{1} p(r)dr .
\end{equation}
AP summarizes a {\it p-r} curve as the weighted mean of precisions achieved at each threshold, with the increase in recall from the previous threshold used as the weight.
Video-specific AP score for evaluating surgical action triplet recognition is computed as follows:
\begin{enumerate}[i.]
    \item \textit{per-category AP} is computed across all frames in a given video.
    \item \textit{category AP} is obtained by averaging per-category APs across all videos.
    \item \textit{mean AP} is obtained by averaging $N$ category AP, serving as the final score: 
\end{enumerate}
\begin{equation}
   mAP = \frac{1}{N} \sum_{i=1}^N AP_i .
\end{equation}
The same process is followed when computing mAP for the individual components of the triplets.
The predictive capacity of a model at recognizing correctly a triplet and its components is evaluated in two ways:
\begin{enumerate}
    \item \textbf{Component average precision:}
    This includes three APs assessing the correct recognition of the instrument ($AP_I$), verb ($AP_V$), and target ($AP_T$) components of the triplets.
    
    \item \textbf{Triplet average precision:}
    This includes three APs assessing the correct recognition of tool-tissue interactions by observing different sets of triplet components, which includes: APs for instrument-verb ($AP_{IV}$), instrument-target ($AP_{IT}$), and instrument-verb-target ($AP_{IVT}$), which is the main metric. 
\end{enumerate}

\subsection{Disentangling Action Triplet Prediction} \label{disentagling_function}
This is introduced to support the evaluation of all triplet component prediction for models that produce only the final triplet ($Y_{IVT}$) probabilities.
If $IVT$ represents all the triplets in a single image, and $D = \{I, V, T, IV, IT\}$ is a set of the triplets' components and their possible combinations, then, $\forall d \in D$, the indirect component's output vector $Y_d$ of class size $C_d$ (e.g. for $d=I, C_d=6$ as there are 6 instrument classes) can be filtered from $Y_{IVT}$ following Equation \ref{eqn:filter}: 

\begin{equation}
    \label{eqn:filter}
    Y_{d} = \biggl[~\max_{d^k \in \{i,v,t\}}~ {Y}_{ivt} ~\biggl]\quad\forall ~ ivt \in IVT, ~ k: 0\leq k < C_d.
\end{equation}

This filtering algorithm \cite{nwoye2021rendezvous} directly translates to obtaining the probability of a given component class as the maximum probability value among all triplet labels having the same component class label in a video frame. For instance, $Y_{hook} = \textnormal{max}(Y_{\triplet{hook, verb, target}}$), $Y_{dissect} = \textnormal{max}(Y_{\triplet{instrument, dissect, target}}$), $Y_{liver} = \textnormal{max}(Y_{\triplet{instrument, verb, liver}}$), etc.
Likewise, the groundtruths of the component labels can be obtained using the same filtering setup.

\subsection{Detection Average Precision}
When evaluating the localization of the triplets as in CholecTriplet2022 challenge, AP metrics consider the overlap or Intersection of Union (IoU) of the predicted bounding boxes with the ground truth. The detection AP can be evaluated in three ways:
\begin{enumerate}
    \item \textbf{Instrument Localization AP}: In this metric, a detection is assigned a true positive (TP) if the degree of overlap (measured as Intersection over Union IoU) between a predicted bounding box (\^b) and the ground truth bounding box (b) of an instrument (I) exceeds a certain threshold $\theta$ (usually 0.5) and the predicted instrument identity (\^y) is correct with respect to the ground truth label (y).
    \begin{equation}
        TP =  (\hat{y}_{_{I}} == y_{_{I}}) + \frac{\hat{b_I} \cap b_I}{\hat{b_I} \cup b_I} \geq \theta.
    \end{equation}
    \item \textbf{Target Localization AP}: Similarly, this metric focus on the correctness of target identification and its bounding box overlap with the groundtruth.
    \begin{equation}
        TP =  (\hat{y}_{_{T}} == y_{_{T}}) + \frac{\hat{b_T} \cap b_T}{\hat{b_T} \cup b_T} \geq \theta
    \end{equation}
    This metric is not yet applicable to the CholecT45 and CholecT50 datasets due to the unavailability of spatial annotations for the targets.
    \item \textbf{Triplet Detection AP}: This metric assess the correctness of every associated action triplet to every localized instrument [and target]. Here, a prediction is considered a TP if the predicted triplet ID is correct, assigned to the right instrument [and target] involved in the tool-tissue interaction, which must also be localized at a minimum IoU threshold with the ground-truth bounding box(es).
    \begin{equation}
        TP =  (\hat{y}_{_{IVT}} == y_{_{IVT}}) + \frac{\hat{b_I} \cap b_I}{\hat{b_I} \cup b_I} \geq \theta ~~ \left [~+~ \frac{\hat{b_T} \cap b_T}{\hat{b_T} \cup b_T} \geq \theta~\right]
    \end{equation}
    
    In future, when {\it target localization AP} is considered, the triplet detection AP will take into account a satisfied bounding box IoU for both instruments and targets. For the meantime, the target localization part is excluded when computing this metric on the datasets.
    
\end{enumerate}
The missed predictions are marked as false negatives (FN) whereas false alarms are marked as false positives (FP).
Following this, their corresponding precision ($p$) and recalls ($r$) are calculated as follows:
\begin{equation}
    \label{metrics:pr}
    \begin{split}
        p &= \frac{TP}{TP+FP}~,\\
        r &= \frac{TP}{TP+FN}~,\\
    \end{split}
\end{equation}
and using the computed $p, r$, the AP is calculated following Equation \ref{metrics:AP}, averaged across videos.

\subsection{Triplet Association Scores (TAS)}
TAS metrics evaluate the quality of a model in associating the bounding box spatial localization with its correct triplet identity. Presently in the CholecT50 dataset, the bounding box localization is on the instrument tips and may in the future consider also the underlying targets.
The triplet association scores are evaluated as follows:
\begin{enumerate}
    \item \textbf{Localize and Match (LM)}: This measures the percentage of the triplets that are correctly predicted and localized at the given overlapping threshold with the groundtruth. The LM considers only the true positive (TP) cases.

    \item \textbf{Partially Localize and Match ($p$LM)}. This computes the percentage of the triplets that are predicted but whose localization overlap with the groundtruth bounding box is less than the considered threshold. 

    \item \textbf{Identity Switch (IDS)}: This calculates the percentage of the triplets that are localized at the given threshold but whose identities are swapped (with other triplets) within the same frame. 

    \item \textbf{Identity Miss (IDM)}: This records the percentage of the triplets that are localized at the given threshold but with an incorrect identity that also does not match any other triplet in the same frame. 

    \item \textbf{Miss Localization (MIL)}: This calculates the percentage of the triplets that are correctly predicted but without a corresponding localization. The MIL metric is useful in evaluating the association capacity of models with parallel recognition and localization branches. 
    
    \item \textbf{Remaining False Positive (RFP)}: This estimates the percentage of false alarms after other factors (i.e. LM, pLM, IDS, IDM, and MIL) have been considered. 
    
    \item \textbf{Remaining False Negative (RFN)}: This estimates the percentage of missed prediction after other factors (i.e. LM, pLM, IDS, IDM, and MIL) have been taken into consideration. 
\end{enumerate}

The TAS metrics are useful in analyzing the capacity of a model in understanding the relationship between presence detection and spatial localization of the triplets. It reveals the usefulness of the learned features for triplet predictions.

The TAS metrics are each expressed in terms of their percentage;
for $ X \in$ \{LM, pLM, IDS, IDM, MIL, RFP, RFN\}:
\begin{equation}
        X_j(\%) = \frac{X_j}{\sum^N_{i=0}{X_i}} \times 100 ,
\end{equation}

to explain the strength and behavior of a model on joint recognition and localization of surgical action triplets.
The TAS metrics is used in CholecTriplet2022 \cite{nwoye2023cholectriplet2022} challenge to provide detailed assessment of the models performance on surgical action triplet detection.

\subsection{Metrics Library}
To standardize the use of these evaluation metrics, we develop {\it ivtmetrics} library, which can be used in both training and inference mode.
The library can be imported in a python-based script using \path{import ivtmetrics} with a prerequisite installation step: \path{pip install ivtmetrics} or \path{conda install -c nwoye ivtmetrics}.
The library provides metrics classes for triplet recognition:
\path{AP = ivtmetrics.Recognition(N : int)} and triplet detection \path{AP = ivtmetrics.Detection(N: int)}, as well as an internally implemented triplet component filtering \path{AP = ivtmetrics.Disentangle(N: int)}, where N = number of triplet classes.
The Detection class inherently computes the triplet association scores based on the TAS metrics.
Invoking the metrics class initializes the metrics accumulators by an \path{AP.reset()} call. This reset function is to be called at the beginning of every training epoch.
Other reset options include \path{AP.reset_video()} to reset scores accumulated for all seen videos, and \path{AP.reset_global()} to reset every accumulator.
The metrics update function takes in the predicted and target labels over each iteration by calling \path{AP.update(targets: array, predictions: array)}. If a video-specific AP is needed, \path{AP.video_end()} must be called at the end of each video.
The AP scores from a current time up to the last \path{reset()} call are obtained via \path{AP.compute_AP(component : str)}. The mean AP average across videos is obtained by calling \path{AP.compute_video_AP(component : str)} while \path{AP.compute_global_AP(component : str)} gives the mean AP across all frames in all seen videos. The component $\in$ ("i", "v", "t", "iv", "it", "ivt") is a string argument that describes the respective sub-task's (instrument, verb, target, instrument-verb, instrument-target, instrument-verb-target) performance to be computed for.
A top $K$ performance is obtained by \path{AP.topk(k: int)} while top predicted class IDs are given by \path{AP.topClass(k: int)}.
The computed results are provided as a dictionary of values with the individual metric names as the keys.
More details and usage examples of the {ivtmetrics} can be found on GitHub \url{https://github.com/CAMMA-public/ivtmetrics} .\\[0.1in]

\section{Benchmark Study Design}
\begin{figure*}[t]
\includegraphics[width=1.0\linewidth]{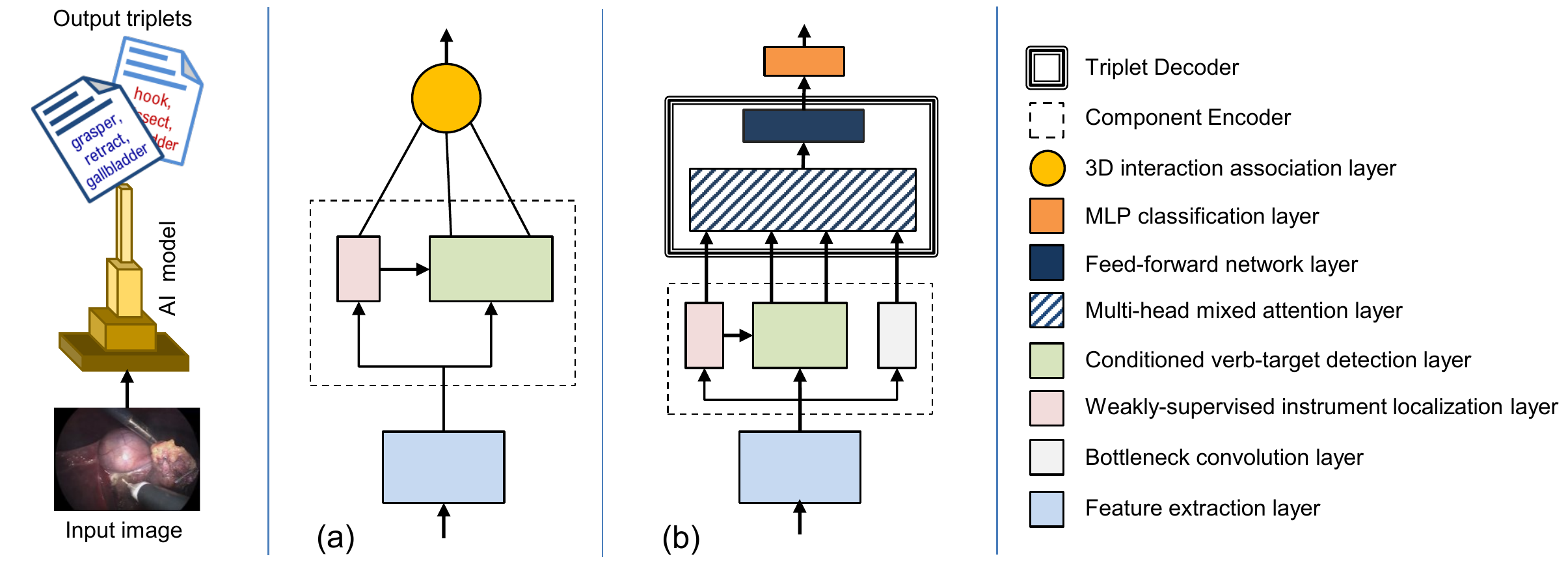}
\caption{Reproduced models for surgical action triplet recognition: (a) Tripnet \cite{nwoye2020recognition}, (b) Rendezvous \cite{nwoye2021rendezvous}.}
\label{fig:methods}
\end{figure*}

To provide a benchmark study on CholecT45 and CholecT50 datasets using our proposed data splits and metrics, we re-implement and reproduce three proposed methods in PyTorch and TensorFlow. 

\subsection{Methods}
We summarize the reproduced methods as follows:
\begin{enumerate}
    \item {\bf Tripnet}: As shown in Fig. \ref{fig:methods}(a), Tripnet \cite{nwoye2020recognition} is a multi-task learning method that uses activation maps resulting from the instrument branch to enhance verb and target feature encoding in a new module known as class activation guide (CAG). It is followed by a 3D interaction space where relationships between instruments-verbs-targets components are resolved to triplets. Code and weights are publicly released on GitHub \url{https://github.com/CAMMA-public/tripnet}.
    
    \item {\bf Attention Tripnet}: This is an upgrade of Tripnet with an attention mechanism. The main difference is the use of class activation guided attention mechanism (CAGAM) \cite{nwoye2021rendezvous} over CAG where the verb and target feature discovery are obtained by channel and position attention processes respectively. Code and weights are publicly released on GitHub \url{https://github.com/CAMMA-public/attention-tripnet}.
    
    \item {\bf Rendezvous (RDV)}: In this model \cite{nwoye2021rendezvous}, the network encoder uses a weakly supervised approach to localize the instruments and the CAGAM module to detect the verb and target components of the triplets as shown in Fig. \ref{fig:methods}(b). The association part is achieved by both self and cross attention mechanisms in a new module known as multi-head of mixed attention (MHMA), and terminated by a simple classifier after 8 successive layers of association decoding. Code and weights are publicly released on GitHub \url{https://github.com/CAMMA-public/rendezvous}.\\[0.1in]
\end{enumerate}

\subsection{Implementation Details}
We made few changes in the original implementations \cite{nwoye2020recognition,nwoye2021rendezvous} as follows:
\begin{enumerate}
    \item {\bf Output resolution}: The original implementation lowered the strides of the last two blocks of the ResNet by one pixel to provide higher resolution ($32 \times 56$) output. However, using original strides of size 2, we implement a faster version (size: $8 \times 14$), trading-off precise localization to speed.
    \item {\bf Attention normalization}: The huge parameter layer-norms in the AddNorm layers of the attention module are replaced with batch-norms without affecting model accuracy.
    \item {\bf Loss function}: We integrates warmup parameters within the auxilliary task's cross-entropies without requiring additional uncertainty loss balancing \cite{kendall2018multi} as in \cite{nwoye2021rendezvous}. This removes the excessive parameters introduced by the uncertainty loss.
\end{enumerate}
\section{Benchmark Results and Discussion}


\begin{table}[ht]
\centering
    \setlength{\tabcolsep}{9pt}
    \captionsetup{skip=0pt,singlelinecheck=off,justification=raggedright}
    \caption{Benchmark triplet recognition AP (\%) on CholecT50 dataset for different frameworks using RDV split.}
    \label{tab:results:rdv}
    \resizebox{\textwidth}{!}{%
        \begin{tabular}{@{}llclcrclcr@{}}
            \toprule
            \multicolumn{1}{l}{\multirow{2}{*}{Framework}}&
            \multicolumn{1}{l}{\multirow{2}{*}{Method}}&
            \multicolumn{4}{r}{Component detection~~~~~}&\phantom{abc}&
            \multicolumn{3}{c}{Triplet association}\\ \cmidrule{4-6} \cmidrule{8-10} 
            \multicolumn{2}{l}{} && \multicolumn{1}{l}{$AP_{I}$~~~~} & \multicolumn{1}{c}{~~$AP_{V}$~~} & \multicolumn{1}{c}{~~$AP_{T}$~~} && \multicolumn{1}{c}{$AP_{IV}$~~~} & \multicolumn{1}{c}{~$AP_{IT}$~} & \multicolumn{1}{r@{}}{$AP_{IVT}$} \\ \midrule
            \multirow{3}{*}{TensorFlow} 
            & Tripnet \cite{nwoye2020recognition} &&  {\bf 92.1} & 54.5 & 33.2 && 29.7 & 26.4 & 20.0  \\ 
            & Attention Tripnet \cite{nwoye2021rendezvous} && 92.0 &  {60.2} & {38.5} && {31.1} & {29.8} & {23.4} \\ 
            & Rendezvous (RDV) \cite{nwoye2021rendezvous} && 92.0 & {60.7} & 38.3 && {39.4} & {\bf 36.9} & {\bf 29.9}  \\ \midrule
            \multirow{3}{*}{PyTorch} 
            
            & Tripnet \cite{nwoye2020recognition} && 88.7 & 59.2 & 39.3 && 31.9 & 27.9 & 21.6 \\
            & Attention Tripnet \cite{nwoye2021rendezvous} && 87.9 & 59.7 & 40.6 && 34.2  & 29.0 & 23.2 \\
            
            & Rendezvous \cite{nwoye2021rendezvous} && 89.1 & \bf 62.3 & \bf 43.8 && \bf 40.0 & 35.8 & 29.5 \\ 
            \bottomrule 
        \end{tabular}
    }
\end{table}

\subsection{Quantitative Results on CholecT50 using Rendezvous Split}
All the 100 triplet classes are evaluated in this setup.
The benchmark results on the RDV split is presented in Table \ref{tab:results:rdv}. The results follow the same trend as in the original paper \cite{nwoye2021rendezvous} as it is observed that the Attention Tripnet leverages CAGAM to improve the verb and target detections while Rendezvous utilizes its transformer-inspired MHMA to improve the triplet association performance.
We show the performance across deep learning frameworks in Table \ref{tab:results:rdv}. The PyTorch models approximates the performance of their TensorFlow counterparts (version 1). We observe that these results are comparable in some of the sub tasks.


\subsection{Quantitative Results on CholecT50 using CholecTriplet Split}
The challenge rule excludes the 6 null triplet classes (IDs: 94-99) from evaluation. The results are presented in Table \ref{table:results:t50}.
It is observed that AP$_{IVT}$ is higher than in the RDV split for each model, likely due to the reduced number of triplet classes (94 v 100).
Also, the direct outputs ($Y_D$) for the individual components (i.e.: $Y_I, Y_V, Y_T, Y_{IV}$ or $Y_{IT}$) of the triplets ($Y_{IVT}$) are not provided by the challenge approaches, instead they are filtered from the main triplet predictions ($Y_{IVT}$) following the filtering formula in Section \ref{disentagling_function}.
As shown in Table \ref{table:results:t50}, the AP performances on the filtered prediction are generally lower compared to AP on directly predicted probabilities of those components when provided, however, they are more informative and a better representation of how a model understands the triplet's composition.


\begin{table}[htp]
\centering
    \setlength{\tabcolsep}{9pt}
    \captionsetup{skip=0pt,singlelinecheck=off,justification=raggedright}
    \caption{Benchmark triplet recognition AP (\%) on CholecT50 dataset using CholecTriplet split.}
    \label{table:results:t50}
    \resizebox{\textwidth}{!}{%
        \begin{tabular}{@{}llclcrclcr@{}}
            \toprule
            \multicolumn{1}{l}{\multirow{2}{*}{Framework}}&
            \multicolumn{1}{l}{\multirow{2}{*}{Method}}&
            \multicolumn{4}{r}{Component detection~~~~~}&\phantom{abc}&
            \multicolumn{3}{c}{Triplet association}\\ \cmidrule{4-6} \cmidrule{8-10} 
            \multicolumn{2}{l}{} && \multicolumn{1}{l}{$AP_{I}$~~~~} & \multicolumn{1}{c}{~~$AP_{V}$~~} & \multicolumn{1}{c}{~~$AP_{T}$~~} && \multicolumn{1}{c}{$AP_{IV}$~~~} & \multicolumn{1}{c}{~$AP_{IT}$~} & \multicolumn{1}{r@{}}{$AP_{IVT}$}  \\ \midrule
            \multirow{3}{*}{TensorFlow} 
            & Tripnet \cite{nwoye2020recognition} && 74.6 &	42.9 & 32.2 &&	27.0 & 28.0 & 23.4 \\
            & Attention Tripnet \cite{nwoye2021rendezvous} && 77.1 & 43.4 & 30.0 && 32.3 & 29.7 & 25.5 \\ 
            & Rendezvous \cite{nwoye2021rendezvous} && \bf 77.5	& \bf 47.5 & 37.7 && 34.4 & \bf 38.2 & {32.7} \\
            \midrule
            \multirow{3}{*}{PyTorch} 
            & Tripnet \cite{nwoye2020recognition} && 73.9 & 41.6 & 32.1 && 28.8 & 29.2 & 27.4 \\ 
            & Attention Tripnet  \cite{nwoye2021rendezvous} && 73.7 & 43.7 & 35.3 && 31.6 & 31.9 & 27.7 \\
            & Rendezvous  \cite{nwoye2021rendezvous} && 75.9 & 46.0 & \bf 38.7 && \bf 35.9 & 37.1 & \bf 32.8 \\
            \bottomrule 
        \end{tabular}
    }
\end{table}


\subsection{Quantitative Results on CholecT50 using Cross-Validation Split}
The benchmarking results on the CholecT50 cross-validation split are presented in Table \ref{table:results:cv50} along with the standard deviation (std) over the folds.
All the 100 triplet classes are evaluated in this setup.
This presents a less biased and less optimistic estimate of the models with confidence intervals.
Their standard deviation (std) spread shows the extent of performances approximation of the three models positioning  Tripnet as the least and Rendezvous as the best in terms of performance.

\begin{table}[ht]
\centering
    \setlength{\tabcolsep}{9pt}
    \captionsetup{skip=0pt,singlelinecheck=off,justification=raggedright}
    \caption{Benchmark triplet recognition AP (\%) on CholecT50 dataset using the official cross-validation split.}
    \label{table:results:cv50}
    \resizebox{\textwidth}{!}{%
        \begin{tabular}{@{}lrlcrrlcr@{}}
            \toprule
            \multirow{2}{*}{Method (in PyTorch)} & \phantom{abc}
            &\multicolumn{3}{c}{{Component detection}} & \phantom{abc}
            &\multicolumn{3}{c}{{Triplet association}} \\ \cmidrule{3-5} \cmidrule{7-9}
            && \multicolumn{1}{c}{$AP_{I}$} & \multicolumn{1}{c}{$AP_{V}$} & \multicolumn{1}{c}{$AP_{T}$} && \multicolumn{1}{c}{$AP_{IV}$} & \multicolumn{1}{c}{$AP_{IT}$} & \multicolumn{1}{c}{$AP_{IVT}$}
            \\ \midrule   
            Tripnet \cite{nwoye2020recognition} &&  89.1±1.7 & 58.8±3.1 & 38.4±1.3 && 32.7±2.4 & 29.0±0.8 & 25.3±2.4 \\
            Attention Tripnet \cite{nwoye2021rendezvous}  && 88.7±1.3 & \textbf{61.1±2.0} & \textbf{40.7±3.2} && 33.1±2.7 & 30.3±1.6	& 27.2±2.9 \\
            Rendezvous \cite{nwoye2021rendezvous}  && \textbf{89.4±2.0} & 60.4±2.8 & 40.3±2.2  &&  \textbf{34.5±2.8} & \textbf{31.8±1.0} & \textbf{29.4±2.5} \\
            \bottomrule
        \end{tabular}
    }
\end{table}


\subsection{Quantitative Results on CholecT45 using Cross-Validation Split}
Similarly, the benchmarking results on the CholecT45 cross-validation split, presented in Table \ref{table:results:cv45}, justifies the use of attention mechanisms for surgical action triplet recognition. The analysis shows that the results obtained on the CholecT45 CV approximates the ones of the CholecT50 CV in all the sub-tasks, justifying its use/sufficiency in the absence of the complete CholecT50 dataset.

\begin{table*}[ht]
\centering
    \setlength{\tabcolsep}{9pt}
    \captionsetup{skip=0pt,singlelinecheck=off,justification=raggedright}
    \caption{Benchmark triplet recognition AP (\%) on CholecT45 dataset using the official cross-validation split.}
    \label{table:results:cv45}
    \resizebox{\textwidth}{!}{%
        \begin{tabular}{@{}lrlcrrlcr@{}}
            \toprule
            \multirow{2}{*}{Method (in PyTorch)} & \phantom{abc}
            &\multicolumn{3}{c}{{Component detection}} & \phantom{abc}
            &\multicolumn{3}{c}{{Triplet association}} \\ \cmidrule{3-5} \cmidrule{7-9}
            && \multicolumn{1}{c}{$AP_{I}$} & \multicolumn{1}{c}{$AP_{V}$} & \multicolumn{1}{c}{$AP_{T}$} && \multicolumn{1}{c}{$AP_{IV}$} & \multicolumn{1}{c}{$AP_{IT}$} & \multicolumn{1}{c}{$AP_{IVT}$}
            \\ \midrule   
            Tripnet \cite{nwoye2020recognition} &&  \textbf{89.9±1.0} & 59.9±0.9 & 37.4±1.5  && 31.8±4.1 & 27.1±2.8 & 24.4±4.7 \\
            Attention Tripnet \cite{nwoye2021rendezvous} && 89.1±2.1 & 61.2±0.6 & {\bf 40.3±1.2} && 33.0±2.9 & 29.4±1.2 & 27.2±2.7 \\
            Rendezvous \cite{nwoye2021rendezvous}  && 89.3±2.1 & \textbf{62.0±1.3} & 40.0±1.4  &&  \textbf{34.0±3.3} & \textbf{30.8±2.1} & \textbf{29.4±2.8} \\
            \bottomrule
        \end{tabular}
    }
\end{table*}

\subsection{Class-wise Performances of the Benchmark Models}
We present the per-class performance for the triplet components (Tables \ref{table:results:instruments} - \ref{table:results:targets}) and their association (Table \ref{table:results:triplets}) recognition using the cross-validation dataset splitting strategy on both CholecT45 and CholecT50. We observe similar performance pattern across the two datasets in all classes of each sub task showing the reliability of cross-validation split approach in model evaluation.

On instrument presence detection, as shown in Table \ref{table:results:instruments}, grasper and hook are the most detected owing to their highest occurrence frequencies in the dataset. However, hook is a little better detected than grasper owing to its uniqueness unlike the grasper which sometimes share some similarities with clipper, bipolar and scissor. The scissors and irrigator, on the other hand, are the least detected likely due to their low occurrence distributions in the datasets.

\begin{table}[ht]
    \captionsetup{skip=0pt,singlelinecheck=off,justification=raggedright}
    \caption{Per-class instrument presence detection AP (\%) on cross-validation splits (Method in PyTorch)}
    \label{table:results:instruments}
    \centering
    \setlength{\tabcolsep}{11pt}
    \resizebox{\textwidth}{!}{%
    \begin{tabular}{@{}lrlcrrlcr@{}}
        \toprule
        \multirow{2}{*}{Classes} & \phantom{abc} &
        \multicolumn{3}{c}{CholecT45} & \phantom{abc} &
        \multicolumn{3}{c}{CholecT50} \\ \cmidrule{3-5} \cmidrule{7-9}
        & & Tripnet & \makecell[r]{Attention\\Tripnet} & RDV && Tripnet & \makecell[c]{Attention\\Tripnet} & RDV \\
        \midrule
        
        grasper && 96.5±0.4 & 96.4±0.7 & 96.6±0.6 && 96.4±0.7 & 96.5±0.6 & 96.6±0.6 \\
        bipolar && 88.4±4.2 & 86.0±4.2 & 87.4±4.7 && 89.0±3.4 & 88.2±4.2 & 88.5±4.1 \\
        hook && 97.5±1.6 & 97.1±1.3 & 97.4±1.5 && 97.5±1.3 & 97.3±1.3 & 97.6±1.3 \\
        scissors && 80.3±6.0 & 79.6±8.4 & 78.4±5.4 && 82.8±4.9 & 81.3±5.8 & 82.6±6.5 \\
        clipper && 91.2±3.9 & 90.1±3.9 & 90.9±3.8 && 89.6±5.6 & 89.8±5.8 & 89.9±5.0 \\
        irrigator && 86.0±4.1 & 85.3±2.8 & 84.5±6.8 && 79.6±6.5 & 79.1±5.7 & 81.3±4.9 \\
        \midrule
        Mean && \bf 89.9±1.0 & 89.1±2.1 & 89.3±2.1 && 89.1±1.7 & 88.7±1.3 & \bf 89.4±2.0 \\
         \bottomrule
    \end{tabular}
    }
\end{table}

For the verb recognition, grasp, retract, dissect, coagulate, clip, cut, and aspirate are better detect above 50\% at all time as shown in Table \ref{table:results:verbs}. This is due to their strong affinities with unique instrument classes. Pack and irrigate are very challenging to discriminate from the dominant verbs of their instruments namely retract and aspirate. Null action, being a compendium of unconsidered actions, is the least recognized verb.

\begin{table}[ht]
    \captionsetup{skip=0pt,singlelinecheck=off,justification=raggedright}
    \caption{Per-class verb recognition AP (\%) on cross-validation splits (Method in PyTorch)}
    \label{table:results:verbs}
    \centering
    \captionsetup{skip=0pt,singlelinecheck=off,justification=raggedright}
    \setlength{\tabcolsep}{9pt}
    \resizebox{\textwidth}{!}{%
    \begin{tabular}{@{}llrlcrrlcr@{}}
        \toprule
        \multirow{2}{*}{Classes} & \phantom{abc}& \phantom{abc} &
        \multicolumn{3}{c}{CholecT45} & \phantom{abc} &
        \multicolumn{3}{c}{CholecT50} \\ \cmidrule{4-6} \cmidrule{8-10}
        & & & Tripnet & \makecell[c]{Attention\\Tripnet} & RDV && Tripnet & \makecell[c]{Attention\\Tripnet} & RDV \\
        \midrule
        grasp &&& 70.5±5.8 & 60.5±9.9 & 69.8±3.7 && 67.1±3.4 & 66.1±5.4 & 68.3±3.0 \\
        retract &&& 90.5±5.4 & 84.0±9.8 & 89.7±7.2 && 86.7±5.1 & 85.8±5.4 & 86.7±5.8 \\
        dissect &&& 93.0±2.8 & 86.5±9.9 & 93.2±3.9 && 90.9±2.4 & 90.6±2.4 & 91.0±3.3 \\
        coagulate &&& 67.2±6.1 & 56.5±9.9 & 68.7±5.5 && 67.9±5.0 & 68.5±6.2 & 69.7±6.1 \\
        clip &&& 85.4±6.4 & 67.8±9.8 & 85.5±3.7 && 85.5±6.3 & 86.1±5.4 & 86.5±5.5 \\
        cut &&& 70.5±9.1 & 57.7±9.9 & 72.0±4.8 && 74.9±3.4 & 72.3±6.1 & 74.9±7.6 \\
        aspirate &&& 60.7±9.2 & 47.1±9.9 & 57.8±9.9 && 57.4±4.9 & 57.1±7.3 & 56.7±5.5 \\
        irrigate &&& 29.6±8.2 & 17.4±9.7 & 25.7±5.8 && 27.6±9.4 & 25.4±7.9 & 25.1±9.2 \\
        pack &&& 32.1±9.9 & 25.8±9.9 & 31.2±9.9 && 26.8±9.9 & 33.2±9.9 & 20.0±9.9 \\
        null-verb &&& 23.0±2.4 & 21.1±5.0 & 24.0±4.1 && 24.5±1.8 & 25.5±4.0 & 24.9±2.6 \\
        \midrule
        Mean &&& 59.9±0.9 & 61.2±0.6 & \bf 62.0±1.3 && 58.8±3.1 & \bf 61.1±2.0 & 60.4±2.8 \\
         \bottomrule
    \end{tabular}
    }
\end{table}

The per-class target detection reveals the most interesting areas of improvement. The predominant targets such as gallbladder, liver, and specimen-bag are well detected above 70\% as shown in Table \ref{table:results:targets}. The main challenge comes in detecting tiny anatomical structures such as cystic-artery, peritoneum, cystic-plate, cytic-pedicle, etc. as against conspicuous structures such as liver, omentum, cystic-duct, fluid, etc. Some anatomies with no clear boundaries such as abdominal wall, cystic-artery, etc. are fairly detected.

\begin{table}[ht]
    \captionsetup{skip=0pt,singlelinecheck=off,justification=raggedright}
    \caption{Per-class target recognition AP (\%) on cross-validation splits (Method in PyTorch)}
    \label{table:results:targets}
    \centering
    \captionsetup{skip=0pt,singlelinecheck=off,justification=raggedright}
    \setlength{\tabcolsep}{7.5pt}
    \resizebox{\textwidth}{!}{%
    \begin{tabular}{@{}lrlcrllcr@{}}
        \toprule
        \multirow{2}{*}{Classes} & \phantom{abc} &
        \multicolumn{3}{c}{CholecT45} & \phantom{abc} &
        \multicolumn{3}{c}{CholecT50} \\ \cmidrule{3-5} \cmidrule{7-9}
        & & Tripnet & \makecell[c]{Attention\\Tripnet} & RDV && Tripnet & \makecell[c]{Attention\\Tripnet} & RDV \\
        \midrule
        gallbladder && 93.6±1.1 & 91.2±6.5 & 93.7±1.2 && 93.6±1.3 & 93.8±1.0 & 93.6±1.6 \\
        cystic-plate && 11.6±3.9 & 10.1±2.0 & 11.0±3.5 && 11.1±2.2 & 11.2±2.7 & 09.9±3.3 \\
        cystic-duct && 47.2±5.8 & 41.9±9.9 & 47.1±2.8 && 47.6±5.6 & 48.1±3.1 & 47.2±3.9 \\
        cystic-artery && 31.9±3.7 & 29.6±9.7 & 31.2±2.2 && 35.0±4.7 & 34.6±2.8 & 35.6±4.6 \\
        cystic-pedicle && 04.0±2.4 & 08.7±6.0 & 13.4±7.8 && 10.3±5.0 & 06.9±8.2 & 10.4±6.5 \\
        blood-vessel && 08.4±5.6 & 15.6±9.9 & 06.7±6.2 && 12.7±9.9 & 23.5±9.9 & 18.7±9.9 \\
        fluid && 58.4±9.2 & 48.9±9.9 & 58.0±9.9 && 56.3±5.1 & 54.5±6.8 & 57.0±5.1 \\
        abdominal-wall/cavity && 30.0±4.6 & 20.4±9.9 & 25.9±7.2 && 25.7±3.4 & 28.5±5.0 & 31.3±9.9 \\
        liver && 71.8±5.5 & 65.3±9.9 & 72.9±2.5 && 72.7±6.5 & 73.5±4.4 & 74.8±4.2 \\
        adhesion && 04.2±0.3 & 13.9±3.3 & 07.2±0.5 && 05.2±0.0 & 33.3±9.9 & 11.3±2.8 \\
        omentum && 46.7±8.6 & 44.4±9.9 & 48.0±9.9 && 45.8±8.0 & 45.8±9.9 & 46.2±9.9 \\
        peritoneum && 17.7±5.5 & 24.1±9.9 & 26.6±4.5 && 19.5±9.9 & 28.8±9.9 & 25.7±8.1 \\
        gut && 10.7±7.7 & 09.6±6.9 & 09.5±6.9 && 13.6±8.3 & 14.4±6.6 & 15.5±7.4 \\
        specimen-bag && 85.8±2.5 & 70.1±9.9 & 84.4±1.2 && 85.5±1.8 & 84.1±1.3 & 84.6±1.2 \\
        null-target && 22.8±2.3 & 21.1±5.4 & 23.5±4.1 && 24.5±2.1 & 25.5±3.9 & 25.2±2.6 \\
        \midrule
        Mean & & 37.4±1.4 & \bf 40.3±1.2 & 40.0±1.4 && 38.4±1.3 & \bf 40.7±3.2 & 40.3±2.2\\
         \bottomrule
    \end{tabular}
    }
\end{table}

A presentation of class-wise results of the complete 100 triplet classes is only possible using the cross-validation approach. 
As shown in Table \ref{table:results:triplets}, the models recognizes the most important triplets such as grasper retracting gallbladder or grasping specimen-bag, hook dissecting either gallbladder or omentum, bipolar coagulating liver, clipper clipping cystic-artery and -duct with scissors cutting the same, and irrigator aspirating fluid or irrigating the cystic-duct. 
The good performance on these specific triplets are expected from the models' high performance on their specific triplet component classes.
Surprising, the irrigator rare use in dissecting cystic-pedicle is well detected by RDV model.

\begin{table}[htp]
    \captionsetup{skip=0pt,singlelinecheck=off,justification=raggedright}
    \caption{Per-class triplet recognition AP (\%) on cross-validation splits (Method in PyTorch)}
    \label{table:results:triplets}
    \centering
    \captionsetup{skip=0pt,singlelinecheck=off,justification=raggedright}
    \setlength{\tabcolsep}{2pt}
    \resizebox{\textwidth}{!}{%
    \begin{tabular}{@{}lr lcr r lcr r lr lcr r lcr@{}}
        \toprule
        \multirow{2}{*}{Classes} & \phantom{abc} &
        \multicolumn{3}{c}{CholecT45} & \phantom{abc} &
        \multicolumn{3}{c}{CholecT50} & \phantom{abc} &
        \multicolumn{1}{c}{\multirow{2}{*}{classes}} & \phantom{abc} &
        \multicolumn{3}{c}{CholecT45} & \phantom{abc} &
        \multicolumn{3}{c}{CholecT50}\\ \cmidrule{3-5} \cmidrule{7-9} \cmidrule{13-15} \cmidrule{17-19}
        & & Tripnet & \makecell[c]{Attention\\Tripnet} & RDV && Tripnet  & \makecell[c]{Attention\\Tripnet} & RDV &&&& Tripnet  & \makecell[c]{Attention\\Tripnet} & RDV && Tripnet  & \makecell[c]{Attention\\Tripnet} & RDV\\
        \midrule
        grasper,dissect,cystic-plate && 01.5±0.3 & 01.5±0.4 & 02.2±1.2 && 03.3±3.4 & 01.7±0.8 & 01.9±0.1 && hook,coagulate,cystic-plate && 00.5±0.1 & 00.0±0.0 & 00.3±0.1 && 01.8±0.1 & 00.4±0.1 & 00.4±0.1\\ 
        grasper,dissect,gallbladder && 09.5±9.5 & 04.6±5.6 & 05.0±4.3 && 06.2±9.7 & 07.7±8.3 & 11.6±9.9 && hook,coagulate,gallbladder && 06.1±8.2 & 03.2±1.7 & 05.6±6.8 && 08.6±9.2 & 03.8±2.9 & 06.7±5.8\\ 
        grasper,dissect,omentum && 01.3±1.2 & 05.1±5.4 & 03.4±4.0 && 03.1±3.7 & 03.0±1.8 & 01.8±1.1 && hook,coagulate,liver && 01.9±1.2 & 02.5±2.1 & 07.5±4.8 && 05.7±4.1 & 04.0±2.9 & 06.1±4.2\\ 
        grasper,grasp,cystic-artery && 01.5±0.3 & 01.3±0.3 & 02.0±0.6 && 01.9±1.3 & 01.3±0.3 & 02.3±0.4 && hook,coagulate,omentum && 07.6±7.6 & 10.3±6.1 & 06.3±7.4 && 13.6±9.9 & 04.8±3.1 & 05.0±5.1\\ 
        grasper,grasp,cystic-duct && 07.9±4.9 & 12.8±6.5 & 20.8±9.9 && 08.0±5.8 & 07.4±3.1 & 18.7±9.9 && hook,cut,blood-vessel && 00.0±0.0 & 00.0±0.0 & 00.0±0.0 && 01.6±0.1 & 01.2±0.1 & 01.0±0.1\\ 
        grasper,grasp,cystic-pedicle && 05.2±1.9 & 20.1±0.1 & 02.8±0.8 && 13.5±9.9 & 02.2±0.2 & 24.0±9.9 && hook,cut,peritoneum && 00.0±0.0 & 00.0±0.0 & 00.0±0.0 && 07.3±0.1 & 05.2±0.1 & 03.4±0.1\\ 
        grasper,grasp,cystic-plate && 20.3±9.9 & 21.6±9.9 & 23.8±9.9 && 07.8±3.8 & 06.0±5.6 & 06.3±4.5 && hook,dissect,blood-vessel && 00.7±0.1 & 01.2±0.1 & 00.9±0.1 && 01.4±0.1 & 00.7±0.1 & 00.8±0.1\\ 
        grasper,grasp,gallbladder && 23.8±9.9 & 22.2±9.9 & 30.5±9.9 && 28.6±9.9 & 28.2±9.9 & 29.4±9.9 && hook,dissect,cystic-artery && 20.4±4.9 & 19.7±6.6 & 20.7±4.3 && 25.5±5.6 & 22.7±4.6 & 26.8±6.9\\ 
        grasper,grasp,gut && 00.4±0.1 & 00.9±0.1 & 00.3±0.1 && 02.1±2.3 & 00.7±0.6 & 01.1±0.4 && hook,dissect,cystic-duct && 37.4±4.4 & 38.7±3.6 & 39.1±3.1 && 37.8±5.5 & 42.5±2.9 & 38.6±3.8\\ 
        grasper,grasp,liver && 02.5±2.0 & 02.3±3.1 & 16.9±9.9 && 07.8±7.7 & 02.1±2.0 & 06.2±4.6 && hook,dissect,cystic-plate && 14.4±6.5 & 11.5±5.1 & 18.3±9.9 && 13.6±7.5 & 17.3±9.9 & 14.5±7.7\\ 
        grasper,grasp,omentum && 04.5±5.8 & 06.1±4.2 & 26.7±9.9 && 08.7±6.2 & 03.8±5.8 & 11.4±9.7 && hook,dissect,gallbladder && 78.7±2.6 & 78.3±3.6 & 78.3±2.2 && 77.5±4.4 & 77.8±4.9 & 77.3±4.2\\ 
        grasper,grasp,peritoneum && 09.2±9.9 & 04.2±4.7 & 03.0±2.4 && 12.3±9.9 & 17.0±9.9 & 15.1±9.9 && hook,dissect,omentum && 62.5±9.9 & 65.1±7.0 & 63.9±8.6 && 67.4±9.9 & 66.5±9.9 & 67.2±9.9\\ 
        grasper,grasp,specimen-bag && 85.3±2.3 & 85.7±1.9 & 84.5±1.1 && 85.3±1.3 & 85.2±1.4 & 84.9±1.4 && hook,dissect,peritoneum && 15.1±2.8 & 11.9±8.5 & 27.3±3.8 && 19.8±6.3 & 32.5±7.4 & 26.9±9.0\\ 
        grasper,pack,gallbladder && 30.9±9.9 & 33.9±9.9 & 35.2±9.3 && 28.4±9.9 & 37.2±7.5 & 28.2±6.9 && hook,retract,gallbladder && 14.8±9.9 & 17.0±5.8 & 23.8±9.9 && 17.9±9.9 & 17.0±8.6 & 21.3±9.9\\ 
        grasper,retract,cystic-duct && 26.9±0.1 & 00.0±0.0 & 45.0±0.1 && 24.1±0.1 & 21.9±0.1 & 38.7±0.1 && hook,retract,liver && 05.0±6.5 & 12.1±3.4 & 19.2±9.9 && 06.8±5.5 & 11.3±9.9 & 11.3±7.9\\ 
        grasper,retract,cystic-pedicle && 00.8±0.1 & 02.1±0.1 & 01.4±0.1 && 01.5±0.1 & 01.2±0.1 & 02.0±0.1 && scissors,coagulate,omentum && 00.8±0.1 & 04.0±0.1 & 01.1±0.1 && 03.1±0.1 & 00.8±0.1 & 01.4±0.1\\ 
        grasper,retract,cystic-plate && 16.0±9.9 & 15.9±1.1 & 17.8±1.3 && 14.7±7.1 & 24.7±9.9 & 15.9±8.7 && scissors,cut,adhesion && 07.9±0.1 & 12.4±0.1 & 10.4±0.1 && 06.1±0.1 & 13.8±0.1 & 11.8±0.1\\ 
        grasper,retract,gallbladder && 83.4±7.6 & 86.5±4.4 & 83.9±9.6 && 79.6±6.9 & 78.3±8.2 & 79.2±8.9 && scissors,cut,blood-vessel && 19.1±9.9 & 36.7±9.9 & 33.4±9.9 && 01.9±1.7 & 10.2±1.3 & 37.5±9.9\\ 
        grasper,retract,gut && 08.5±5.2 & 10.5±6.0 & 10.8±5.1 && 18.3±8.9 & 13.7±6.2 & 17.4±9.9 && scissors,cut,cystic-artery && 50.6±9.9 & 62.1±4.8 & 57.3±5.6 && 56.1±5.8 & 58.9±9.4 & 58.9±7.2\\ 
        grasper,retract,liver && 69.7±6.7 & 72.1±6.8 & 72.0±2.6 && 71.0±6.2 & 71.0±4.7 & 74.1±3.9 && scissors,cut,cystic-duct && 51.8±9.9 & 56.3±5.6 & 59.0±7.3 && 56.4±5.8 & 56.7±5.7 & 58.6±4.2\\ 
        grasper,retract,omentum && 44.9±9.9 & 43.0±9.9 & 45.5±9.9 && 42.1±9.9 & 42.6±9.9 & 47.9±9.9 && scissors,cut,cystic-plate && 01.5±1.6 & 16.0±2.5 & 22.9±6.1 && 25.5±9.9 & 09.3±4.3 & 48.5±9.9\\ 
        grasper,retract,peritoneum && 17.7±9.9 & 31.3±9.9 & 43.5±9.9 && 24.0±9.9 & 50.5±9.9 & 46.5±9.9 && scissors,cut,liver && 02.1±0.1 & 14.9±0.1 & 08.8±0.1 && 25.4±0.1 & 06.3±0.1 & 23.9±0.1\\ 
        bipolar,coagulate,abdominal-wall-cavity && 41.2±9.9 & 40.0±9.9 & 35.6±9.9 && 39.4±9.9 & 45.9±8.1 & 41.1±9.9 && scissors,cut,omentum && 01.9±0.1 & 00.0±0.0 & 07.9±0.1 && 04.9±5.5 & 01.7±0.8 & 21.0±9.9\\ 
        bipolar,coagulate,blood-vessel && 05.5±4.1 & 12.2±9.1 & 24.0±9.9 && 23.2±9.9 & 50.8±9.9 & 41.3±9.9 && scissors,cut,peritoneum && 02.7±0.1 & 07.4±0.1 & 42.9±0.1 && 02.8±0.1 & 04.2±0.1 & 23.4±0.1\\ 
        bipolar,coagulate,cystic-artery && 21.3±1.5 & 03.9±0.1 & 15.4±4.9 && 11.3±6.4 & 09.4±3.7 & 26.2±9.9 && scissors,dissect,cystic-plate && 00.4±0.1 & 00.4±0.1 & 02.0±0.1 && 00.5±0.1 & 00.6±0.1 & 00.8±0.1\\ 
        bipolar,coagulate,cystic-duct && 02.6±0.1 & 03.8±0.1 & 07.7±0.1 && 00.8±0.1 & 01.3±0.1 & 01.4±0.1 && scissors,dissect,gallbladder && 02.3±0.1 & 00.0±0.0 & 03.7±0.1 && 02.7±0.1 & 00.9±0.1 & 01.4±0.1\\ 
        bipolar,coagulate,cystic-pedicle && 27.6±9.9 & 36.9±9.9 & 45.5±9.9 && 32.2±9.9 & 32.3±9.9 & 50.0±9.9 && scissors,dissect,omentum && 04.5±0.1 & 15.8±0.1 & 06.6±0.1 && 08.3±0.1 & 04.7±0.1 & 48.4±0.1\\ 
        bipolar,coagulate,cystic-plate && 29.3±9.9 & 25.6±9.9 & 40.5±9.9 && 31.7±9.9 & 35.8±9.9 & 33.3±9.9 && clipper,clip,blood-vessel && 13.6±5.8 & 15.5±9.9 & 17.4±9.9 && 12.1±9.9 & 20.9±9.9 & 24.9±3.6\\ 
        bipolar,coagulate,gallbladder && 36.6±9.9 & 52.4±9.9 & 43.7±9.9 && 43.0±9.9 & 41.3±9.9 & 48.5±9.9 && clipper,clip,cystic-artery && 58.7±4.1 & 61.2±9.9 & 66.5±4.0 && 57.9±9.5 & 61.6±9.9 & 67.4±3.5\\ 
        bipolar,coagulate,liver && 77.6±7.4 & 79.7±5.9 & 78.2±6.8 && 79.8±4.5 & 79.7±7.6 & 80.9±7.8 && clipper,clip,cystic-duct && 65.2±9.3 & 70.0±5.7 & 70.7±6.1 && 68.6±6.4 & 70.6±8.2 & 73.0±4.9\\ 
        bipolar,coagulate,omentum && 33.3±9.9 & 42.8±9.9 & 37.0±9.9 && 49.5±9.9 & 44.7±9.9 & 46.9±9.9 && clipper,clip,cystic-pedicle && 03.5±0.1 & 05.2±0.1 & 26.8±0.1 && 12.5±0.1 & 00.6±0.1 & 00.8±0.1\\ 
        bipolar,coagulate,peritoneum && 08.3±0.1 & 00.0±0.0 & 22.5±0.1 && 10.3±6.6 & 61.7±9.9 & 33.2±9.9 && clipper,clip,cystic-plate && 02.4±0.9 & 12.0±9.3 & 16.3±9.1 && 10.7±8.3 & 16.3±9.9 & 19.9±9.9\\ 
        bipolar,dissect,adhesion && 07.0±0.1 & 00.0±0.0 & 05.1±0.1 && 07.8±0.1 & 02.0±0.1 & 07.9±0.1 && irrigator,aspirate,fluid && 58.9±9.9 & 57.3±3.0 & 57.4±9.9 && 57.7±4.4 & 56.0±5.5 & 58.9±3.6\\ 
        bipolar,dissect,cystic-artery && 07.0±5.1 & 29.8±9.9 & 22.3±9.9 && 15.8±9.9 & 20.3±9.9 & 32.5±9.9 && irrigator,dissect,cystic-duct && 04.7±0.1 & 00.0±0.0 & 18.1±0.1 && 13.3±0.1 & 11.8±0.1 & 04.5±0.1\\ 
        bipolar,dissect,cystic-duct && 25.9±9.9 & 25.9±3.5 & 08.9±4.5 && 07.3±5.7 & 18.0±9.9 & 22.9±9.9 && irrigator,dissect,cystic-pedicle && 18.8±0.5 & 39.5±9.9 & 60.6±9.9 && 36.3±9.9 & 52.0±9.9 & 51.0±5.4\\ 
        bipolar,dissect,cystic-plate && 04.5±1.9 & 08.7±0.1 & 03.5±1.3 && 16.6±9.9 & 05.2±2.7 & 19.1±1.7 && irrigator,dissect,cystic-plate && 01.2±0.1 & 02.5±0.1 & 02.0±0.1 && 00.4±0.1 & 04.3±0.1 & 02.2±0.1\\ 
        bipolar,dissect,gallbladder && 23.0±9.9 & 39.3±9.9 & 20.5±6.3 && 12.2±9.9 & 20.0±9.9 & 19.5±9.9 && irrigator,dissect,gallbladder && 02.3±2.0 & 11.1±7.5 & 19.6±9.9 && 03.0±3.0 & 05.2±6.5 & 11.4±3.8\\ 
        bipolar,dissect,omentum && 11.2±0.1 & 00.0±0.0 & 26.0±0.1 && 09.2±0.1 & 53.2±0.1 & 33.0±0.1 && irrigator,dissect,omentum && 03.4±3.4 & 13.5±9.9 & 08.8±4.1 && 04.2±5.0 & 13.3±9.9 & 11.8±7.7\\ 
        bipolar,grasp,cystic-plate && 00.8±0.1 & 00.3±0.1 & 00.5±0.1 && 00.6±0.1 & 00.3±0.1 & 00.5±0.1 && irrigator,irrigate,abdominal-wall-cavity && 23.0±9.9 & 17.8±9.9 & 28.2±5.8 && 26.3±9.9 & 28.8±6.3 & 25.1±9.9\\ 
        bipolar,grasp,liver && 03.4±0.1 & 95.5±0.1 & 15.2±0.1 && 21.2±0.1 & 99.5±0.1 & 29.8±0.1 && irrigator,irrigate,cystic-pedicle && 92.4±2.6 & 96.1±6.0 & 92.8±2.0 && 94.0±4.6 & 91.1±0.6 & 01.8±0.8\\ 
        bipolar,grasp,specimen-bag && 23.3±9.9 & 25.6±9.9 & 26.7±9.9 && 17.0±9.9 & 23.7±9.9 & 16.6±9.9 && irrigator,irrigate,liver && 13.3±9.9 & 26.9±9.9 & 18.0±8.0 && 21.1±7.9 & 19.5±8.3 & 21.6±7.7\\ 
        bipolar,retract,cystic-duct && 00.2±0.1 & 00.2±0.1 & 01.7±0.1 && 00.2±0.1 & 00.3±0.1 & 01.4±0.1 && irrigator,retract,gallbladder && 19.1±9.9 & 21.9±9.9 & 48.5±9.9 && 42.8±9.9 & 07.1±6.1 & 33.3±9.9\\ 
        bipolar,retract,cystic-pedicle && 00.8±0.1 & 37.6±0.1 & 38.2±0.1 && 00.5±0.1 & 01.4±0.1 & 07.3±0.1 && irrigator,retract,liver && 16.7±3.2 & 24.7±9.9 & 27.5±6.0 && 15.2±6.1 & 21.7±9.9 & 24.5±4.9\\ 
        bipolar,retract,gallbladder && 01.0±0.2 & 01.9±1.5 & 01.7±0.7 && 00.8±0.5 & 03.5±2.8 & 18.0±9.9 && irrigator,retract,omentum && 05.1±6.7 & 02.8±2.7 & 11.0±9.9 && 06.8±7.2 & 23.6±9.9 & 03.2±1.9\\ 
        bipolar,retract,liver && 15.7±9.9 & 11.0±5.4 & 13.3±6.5 && 13.4±4.3 & 12.4±5.2 & 12.9±6.0 && grasper,null-verb,null-target && 22.6±4.8 & 23.0±4.9 & 24.4±5.6 && 24.5±3.2 & 25.2±4.1 & 24.6±3.5\\ 
        bipolar,retract,omentum && 05.6±4.6 & 14.8±6.7 & 17.1±9.4 && 20.4±8.4 & 21.6±9.9 & 17.9±4.2 && bipolar,null-verb,null-target && 13.0±4.0 & 14.8±7.7 & 14.0±7.4 && 16.6±9.9 & 14.5±8.0 & 19.0±9.9\\ 
        hook,coagulate,blood-vessel && 01.2±1.3 & 00.5±0.1 & 01.7±1.4 && 01.0±0.6 & 01.9±2.7 & 01.6±1.8 && hook,null-verb,null-target && 15.8±4.6 & 17.5±2.3 & 17.0±2.7 && 17.4±1.9 & 18.0±4.3 & 19.7±1.7\\ 
        hook,coagulate,cystic-artery && 00.5±0.1 & 00.6±0.1 & 01.3±0.1 && 00.8±0.1 & 01.3±0.1 & 00.4±0.1 && scissors,null-verb,null-target && 06.8±2.5 & 23.9±9.9 & 15.1±9.9 && 15.4±9.5 & 18.4±8.7 & 18.2±9.9\\ 
        hook,coagulate,cystic-duct && 02.6±3.7 & 00.5±0.5 & 02.8±1.8 && 01.4±1.1 & 01.6±1.4 & 05.1±7.3 && clipper,null-verb,null-target && 25.4±9.9 & 22.6±9.9 & 33.0±9.9 && 15.9±8.5 & 20.4±8.7 & 24.9±9.9\\ 
        hook,coagulate,cystic-pedicle && 00.9±0.4 & 00.5±0.2 & 05.6±7.5 && 01.3±1.3 & 00.7±0.4 & 00.5±0.1 && irrigator,null-verb,null-target && 15.2±9.0 & 14.7±6.6 & 13.1±3.8 && 16.4±8.6 & 16.8±9.2 & 14.0±8.9 \\
        \midrule
        Mean &&&&&&&&&&&& 24.4±04.7 & 27.2±02.7 & \bf29.4±02.8 && 25.3±02.4 & 27.2±02.9 & \bf 29.4±02.5 \\
         \bottomrule
    \end{tabular}
    }
\end{table}

\section{Conclusion} 
With the first public release of the CholecT45 and dataset to support research on surgical action triplet recognition, we present in this paper a standard practice for splitting the dataset to enable a uniform comparison of research methods. These splits remain relevant in the release of the entire CholecT50 dataset.
We also design, implement, and publicly release a python packaged metrics library, {\it ivtmetrics} for the evaluation of surgical action triplet recognition and localization on the dataset.
For the benchmark study, we re-implement the current state-of-the-art models in two predominant deep learning frameworks, PyTorch and TensorFlow, and then train and evaluate them on the proposed data splits.
Owing to the well-articulated rationale for dataset splits and exhaustive cross-validation, results obtained reflect better the generalization capability of the models.
This study sets a rich foundation for fair comparison of methods researched on the CholecT45 and CholecT50 datasets using the same data splits and metrics.
Future work will extend the metrics library to include more statistical evaluations.

\subsection*{Acknowledgements: }
This work was supported by French state funds managed within the Investissements d`Avenir program by BPI France (project CONDOR) and by the ANR under references ANR-11- LABX-0004 (Labex CAMI), ANR-16-CE33-0009 (DeepSurg), ANR-10-IAHU-02 (IHU Strasbourg) and ANR-20-CHIA-0029-01 (National AI Chair AI4ORSafety). 
It was granted access to the HPC resources of Unistra Mesocentre and GENCI-IDRIS (Grant 2021-AD011011638R1).
The authors also thank the IHU and IRCAD research teams for their help with the initial data annotation during the CONDOR project.

This paper is constantly updated with methods (+ results) that follows the recommended data splits and metrics for surgical action triplet recognition and detection.
\bibliographystyle{IEEEtran}
\bibliography{main} 

\end{document}